\documentclass{Interspeech}

\interspeechcameraready

\title{Analyzing Mitigation Strategies for Catastrophic Forgetting in End-to-End Training of Spoken Language Models}

\author[affiliation={1}]{Chi-Yuan}{Hsiao}
\author[affiliation={1}]{Ke-Han}{Lu}
\author[affiliation={1}]{Kai-Wei}{Chang}
\author[affiliation={1}]{Chih-Kai}{Yang}
\author[affiliation={1}]{Wei-Chih}{Chen}
\author[affiliation={1}]{Hung-yi}{Lee}

\affiliation{}{National Taiwan University}{Taiwan}
\email{r12942086@ntu.edu.tw, d12942024@ntu.edu.tw, kaiwei.chang.tw@gmail.com, chihkaiyang1124@gmail.com, r12921120@ntu.edu.tw, hungyilee@ntu.edu.tw}

\usepackage{comment}
\usepackage{multirow}
\usepackage{relsize}
\usepackage{makecell}
\usepackage{caption}
\begin{document}

\maketitle

    
    
\begin{abstract}

    End-to-end training of Spoken Language Models (SLMs) commonly involves adapting pre-trained text-based Large Language Models (LLMs) to the speech modality through multi-stage training on diverse tasks such as ASR, TTS and spoken question answering (SQA). Although this multi-stage continual learning equips LLMs with both speech understanding and generation capabilities, the substantial differences in task and data distributions across stages can lead to catastrophic forgetting, where previously acquired knowledge is lost. This paper investigates catastrophic forgetting and evaluates three mitigation strategies—model merging, discounting the LoRA scaling factor, and experience replay to balance knowledge retention with new learning. Results show that experience replay is the most effective, with further gains achieved by combining it with other methods. These findings provide insights for developing more robust and efficient SLM training pipelines.

\end{abstract}

\section{Introduction}
Inspired by the remarkable success of large language models (LLMs) \cite{achiam2023gpt, team2023gemini,dubey2024llama, yang2024qwen25} in natural language processing (NLP), researchers have begun exploring \emph{spoken language models} (SLMs)~\footnote{Currently, there is no strict definition of SLMs. In this paper, we define SLMs as models that can process both text and speech as input and generate either text or speech as output, with at least one modality being speech. Without loss of generality, this paper focus on analysing SLMs capable of simultaneously accepting text and speech as input and producing both text and speech as output.} as powerful solutions for speech processing tasks. For instance, textless SLMs~\cite{lakhotia2021generative} perform speech continuation without text supervision, while task-specific SLMs, such as VALL-E~\cite{10842513} for text-to-speech (TTS) and Seamless~\cite{barrault2023seamless} for speech translation, leverage generative language modeling to achieve state-of-the-art performance. More recently, researchers have also investigated the instruction-following (IF) capability of SLMs, enabling them to tackle diverse speech processing tasks through natural language guidance. This advancement enhances the flexibility and adaptability of SLMs across various applications~\cite{huang2024dynamicsuperbphase2collaborativelyexpanding, 10448257, arora2025landscape, arora-etal-2024-universlu, tian-etal-2025-espnet, gong2023joint, tang2024salmonn, lu2024desta, lu2024developing, chu2023qwen, chu2024qwen2}.


Due to the high complexity of speech signals, advanced SLMs are typically built by incorporating pre-trained text LLMs rather than being trained from scratch. A common approach is to use a pre-trained text LLM as the backbone and adapt it to the speech modality, allowing it to understand and/or generate speech~\cite{gong2023joint, tang2024salmonn, lu2024desta, zhang-etal-2023-speechgpt}. 
These models can be broadly categorized based on how they incorporate speech into the LLM. One approach involves integrating a \emph{speech encoder} with a text LLM through a projection network for representation alignment. The projection network are then trained optionally along with the LLM to make it familiar with the speech modality, as seen in models like Qwen-Audio~\cite{chu2023qwen,chu2024qwen2}, SALMONN~\cite{tang2024salmonn}, and DeSTA\cite{lu2024desta,lu2024developing}. While these SLMs demonstrate strong speech understanding capabilities, they cannot generate speech responses.

Another approach directly integrates speech tokens (e.g., semantic tokens derived from self-supervised learning (SSL) speech models and acoustic tokens from speech codec models~\cite{10158503}) into the LLM, as seen in models~\cite{defossez2024moshi, xie2024mini, zeng2024glm, yang2024building} such as Moshi and Mini-Omni. This usually requires \emph{vocabulary expansion}~\cite{zhang-etal-2023-speechgpt, xie2024mini}, where the LLM's vocabulary is extended to include both text and speech tokens. By jointly modeling text and speech tokens, these SLMs can effectively understand and generate both modalities. 

\begin{figure}[t]
  \centering
  \includegraphics[width=\linewidth]{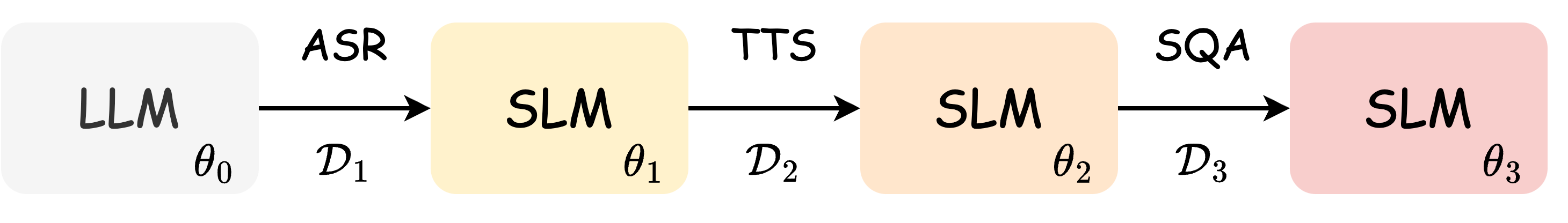}
  \caption{Continual training of a spoken language model using multi-stage speech processing tasks.}
  \label{fig:slm-lifelong}
\end{figure}

In order to familiarize LLMs with the speech modality, \emph{multi-stage training} is often employed. This involves training the LLM on speech processing tasks across several stages, each using a distinct dataset, such as ASR, TTS, and Spoken Question Answering (SQA) as shown in Fig.~\ref{fig:slm-lifelong}. For example, during the ASR stage, the LLM is equipped with speech understanding capabilities, while in the TTS stage, the LLM learns to generate speech. In the SQA stage, the LLM gains the ability to answer questions in speech based on spoken input.

However, due to substantial differences in tasks and data distributions across stages, \emph{catastrophic forgetting}~\cite{goodfellow2013empirical} may occur, causing the LLM to lose previously acquired knowledge or abilities. Although SLMs gain new speech understanding capabilities, they must also retain their original text-based knowledge for tasks such as SQA or following speech instructions. Catastrophic forgetting can degrade the performance of SLMs in both text and speech modalities.

To study this problem and explore potential solutions, we examine three widely used strategies for mitigating catastrophic forgetting in LLMs and SLMs. These strategies include (1) model merging~\cite{yang2024model, lin2024mitigating}, (2) discounting the LoRA scaling factor~\cite{tang2024salmonn, lu24c_interspeech}, and (3) experience replay~\cite{rolnick2019experience, zheng2025lifelong, zhang2023vqacl}. Specifically, in this paper, we train an SLM based on LLaMA\cite{dubey2024llama} and systematically analyze catastrophic forgetting at each training stage by evaluating its performance on question answering and instruction-following tasks in the text modality to assess knowledge retention. After applying mitigation strategies, we compare their effectiveness by evaluating SQA in the speech modality, along with the previously tested text-based tasks.

Overall, this study investigates catastrophic forgetting in SLM training and evaluates three commonly used mitigation strategies. Our experimental results show that among the three strategies examined, experience replay proves to be the most effective, significantly reducing knowledge loss while maintaining performance across both text and speech modalities.


\section{Mitigation strategies}
In this section, we present three common strategies for mitigating catastrophic forgetting in LLMs and SLMs, which are the focus of this paper: (1) model merging~\cite{yang2024model, lin2024mitigating}, (2) discounting the LoRA scaling factor~\cite{tangsalmonn, lu2024desta}, and (3) experience replay~\cite{rolnick2019experience, zheng2025lifelong, zhang2023vqacl}.


\subsection{Model merging}

Consider an SLM training process with \( N \) stages, where \( \theta_i \) denotes the model weights after the \( i \)-th training stage. The complete set of model weights is given by \( M = \{\theta_0, \theta_1, \theta_2, \dots, \theta_N\} \), where \( \theta_0 \) represents the original pre-trained model, and \( \theta_N \) the final SLM weights. To leverage information from multiple training stages, we explore model merging techniques that aggregate weights in \( M \) using several methods, including naive linear combination method, TIES~\cite{TIES}, and DARE~\cite{DARE}. By applying these methods, we aim to preserve knowledge from different training stages, mitigating forgetting and enhancing the final performance.




\subsection{Discounting LoRA-scaling factor}

Given an input \(\mathbf{x} \in \mathbb{R}^{d_{\text{in}}}\) and a weight matrix \(\mathbf{W} \in \mathbb{R}^{d_{\text{out}} \times d_{\text{in}}}\), the forward pass with a LoRA adapter of rank \(r\) is given by:
\begin{align}
    \mathbf{y} = \mathbf{W} \mathbf{x} + \frac{\alpha}{r} (\mathbf{B} \mathbf{A} \mathbf{x}),
\end{align}
where \(\mathbf{A} \in \mathbb{R}^{r \times d_{\text{in}}}\) and \(\mathbf{B} \in \mathbb{R}^{d_{\text{out}} \times r}\) are the weight of the adapter, and \(\alpha\) is the scaling factor. The strategy is to set a lower \(\alpha\) of model that is finetuned with LoRA adapter during inference process to make the effect of adapter weaker.

\subsection{Experience replay}

Different from above strategies, experience replay is applied while model training. Suppose a pre-trained model \(\theta_0\) is initially trained on a dataset \(\mathcal{D}_0\) and will undergo \(N\) additional training stages. We define the set of training datasets as:  
\begin{align}
    D = \left\{\mathcal{D}_1, \mathcal{D}_2, \dots, \mathcal{D}_N\right\},
\end{align}
where \(\mathcal{D}_i\) represents the dataset used in the \(i\)-th training stage.  

At each stage \(i\), experience replay is applied to construct an augmented dataset \(\mathcal{D}_i'\) by including random samples from all previous datasets as well as \(\mathcal{D}_0\), defined as:  
\begin{align}
    \mathcal{D}_i' = \mathcal{D}_i \cup \bigcup_{j=0}^{i-1} \text{Sample}(\mathcal{D}_j, s |\mathcal{D}_i|),
\end{align}
where $Sample(\mathcal{D},k)$ means randomly sample $k$ examples from dataset $D$. \(s\) is the sampling ratio, determining the proportion of each previous dataset \(\mathcal{D}_j\) included in \(\mathcal{D}_i'\). Since each dataset \(\mathcal{D}_i\) may corresponds to different task, the multi-task learning is applied when training with the augmented dataset \(\mathcal{D}_i'\). Notably, the number of random samples is according to the size of i-th dataset \(|\mathcal{D}_i|\).

\section{Experimental setup}
\subsection{Spoken language model}
\begin{figure}[t]
  \centering
  \includegraphics[width=\linewidth]{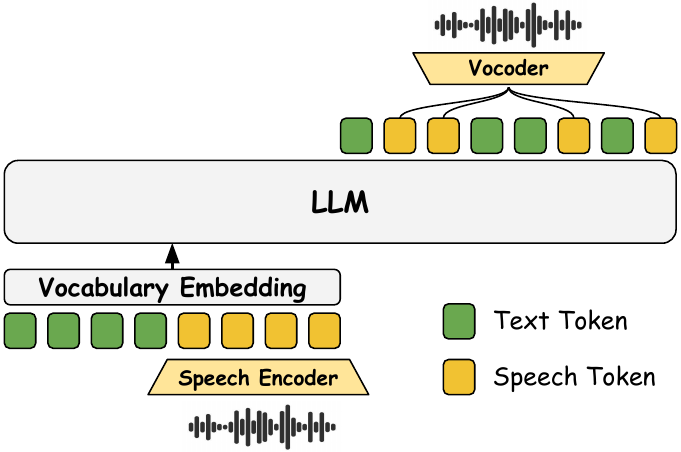}
\caption{The architecture of a Spoken Language Model (SLM), which consists of a backbone LLM, a speech encoder that converts speech into speech tokens, and a vocoder that synthesizes the speech tokens into a speech waveform.}

  \label{fig:slm}
\end{figure}

\subsubsection{Model architecture}
As shown in Figure~\ref{fig:slm}, our SLM comprises three main components: a speech encoder, a LLM backbone, and a vocoder. The speech encoder extracts speech features from speech waveforms, subsequently quantized into discrete speech tokens via k-means clustering. These tokens are incorporated into the LLM’s vocabulary for language modeling. Finally, the vocoder reconstructs speech waveform from the generated speech tokens.
\subsubsection{Training methods}
We fine-tune the LLM in three stages of instruction-tuning for different tasks, including automatic speech recognition (ASR), text-to-speech synthesis (TTS), and spoken question answering (SQA). During fine-tuning, loss is computed only on the model's response. Formally, let $\textbf{P}$ denotes a prompt and $\textbf{R}$ the model's response. 
A text sentence $\textbf{T}$ with $L$ words is defined as $\textbf{T} = [\mathbf{t}_1, \mathbf{t}_2, \mathbf{t}_3, \dots, \mathbf{t}_L]$, where $\mathbf{t}_i$ represents the text token sequence of the $i$-th word.  Similarly, a speech utterance $\textbf{S}$ with $L$ spoken words is defined as $\textbf{S} = [\mathbf{s}_1, \mathbf{s}_2, \mathbf{s}_3, \dots, \mathbf{s}_L]$, where $\mathbf{s}_i$ represents the speech token sequence of the $i$-th word. Each $\mathbf{t}_i$ and $\mathbf{s}_i$ can have varying lengths, containing a different number of text tokens and speech tokens, respectively. We outline the data formulation and methodology for each training stage:


    
\textbf{ASR stage:} To align speech and text tokens, we first train the model on automatic speech recognition (ASR). In this stage, the model learns to generate a text transcription $\textbf{T}_{ASR}$ given a text instruction $\textbf{T}_I$ for ASR and a speech utterance $\textbf{S}_{ASR}$. The prompt $\textbf{P}$ and the model response $\textbf{R}$ are shown as: 
\begin{align}
    \textbf{P} = [\textbf{T}_I, \textbf{S}_{ASR}], \textbf{R} = [\textbf{T}_{ASR}],
\end{align}
where $\textbf{S}_{ASR}$ and $\textbf{T}_{ASR}$ are the speech-text pair in ASR dataset.

\textbf{TTS stage:} Next, the model learns speech generation using the same ASR dataset, as speech-text pairs are also required for text-to-speech synthesis (TTS). In this stage, the model generates text and speech tokens in an alternating, word-by-word interleaved manner—an approach crucial for successful training, as the model often struggles to converge without it. Given a text instruction $\textbf{T}_I$ for TTS and a text sentence $\textbf{T}_{ASR}$. The prompt and model response are: 
\begin{align}
    \textbf{P} = [\textbf{T}_I, \textbf{T}_{ASR}], \textbf{R} = [\mathbf{t}_1, \mathbf{s}_1, \mathbf{t}_2, \mathbf{s}_2, ..., \mathbf{t}_L, \mathbf{s}_L],
\end{align}
where $\mathbf{t}_i \in \textbf{T}_{ASR}, \mathbf{s}_i \in \textbf{S}_{ASR} \forall i=1,2,...,L$, and $\textbf{S}_{ASR}$ and $\textbf{T}_{ASR}$ are from the same speech-text pairs in ASR dataset. 
    
\textbf{SQA stage:} Finally, the model learns spoken question answering (SQA) by leveraging its ASR and TTS capabilities. Given a spoken question $\textbf{S}_{Q}$, the model first predicts its text transcription $\textbf{T}_{Q}$, followed by the text answer $\textbf{T}_{A}$ and its word-by-word interleaved text-speech representation. The prompt and response are structured as:
\begin{align}
    \textbf{P} = [\textbf{S}_{Q}], \textbf{R} = [\textbf{T}_{Q}, \textbf{T}_{A}, \mathbf{t}_1, \mathbf{s}_1, \mathbf{t}_2, \mathbf{s}_2, ..., \mathbf{t}_L, \mathbf{s}_L],
\end{align}
where $\mathbf{t}_i \in \textbf{T}_{A}, \mathbf{s}_i \in \textbf{S}_{A} \forall i=1,2,...,L$. $\textbf{S}_{Q}$, $\textbf{T}_{Q}$, $\textbf{S}_{A}$ and $\textbf{T}_{A}$ originate from the same SQA dataset example.

\textbf{Experience replay:} During training with experience replay, data formulation follows the same structure as in each training stage, including samples from both the current and previous stages. The initial dataset $\mathcal{D}_0$, representing the LLM's original training data, is assumed to contain text instruction-response pairs. If such a dataset is available, the prompt and response are formulated as:
\begin{align}
    \textbf{P} = [\textbf{T}_{I}], \textbf{R} = [\textbf{T}_R],
\end{align}
where $\textbf{T}_{I}$ and $\textbf{T}_{R}$ are the text instruction-response pair in $\mathcal{D}_0$. 

\subsubsection{Training details}
We follow SeamlessM4T v2's settings~\cite{barrault2023seamless} for speech token extraction and reconstruction. Our model uses xlsr2-1b-v2~\cite{conneau21_interspeech} as speech encoder with k-means clustering~\cite{764879} to obtain 10,000 discrete speech tokens. The LLM is based on Llama-3.2-11B-Vision-Instruct, excluding the vision encoder. For vocoding, we adopt the pre-trained HiFi-GAN~\cite{kong2020hifi} from SeamlessM4T v2, supporting multiple speakers and languages. Our hyperparameters include LoRA adapters of rank $r=64$ and $\alpha=16$ for self-attention matrices, with full fine-tuning of the embedding layer, language model heads, and the last five self-attention layers. We optimize the model using the AdamW optimizer with a learning rate of 1e-5 and a warmup ratio 0.1. In ASR and TTS stage, we train the whole dataset for 2 epochs with batch size 4 and set max sequence length to 800. In SQA stage, we train the whole dataset for 1 epoch with batch size 1 and set max sequence length to 1200. All experiments were conducted on four Nvidia RTX A6000 GPUs, with the full training process taking approximately five days.

\subsection{Mitigation strategies}

\subsubsection{Model merging}
Our training process consists of three stages, resulting in a set of model weights denoted as $M = \{\theta_0, \theta_1, \theta_2, \theta_3\}$, where $\theta_0$, $\theta_1$, $\theta_2$, and $\theta_3$ represent the model weights at the initial stage, after the ASR stage, after the TTS stage, and after the SQA stage, respectively. The following introduces the settings for each model merging method:

\textbf{Linear Combination:} $\text{weight} = [0.02, 0.03, 0.05, 0.9]$

\textbf{TIES:} $\text{weight} = [-, 0.04, 0.06, 0.9]$, $\text{density} = [-, 0.9, 0.9, 0.9]$, $\text{BaseModel} = \theta_0$

\textbf{DARE:} $\text{weight} = [-, 0.04, 0.06, 0.9]$, $\text{density} = [-, 0.9, 0.9, 0.9]$, $\text{BaseModel} = \theta_0$




\textbf{Discounting LoRA-scaling factor}
Restricted to computation budget, we only choose $\alpha=15$ and $\alpha=14$ as hyperparameter of LoRA adapters for evaluation.

\textbf{Experience replay:} We evaluate two training settings: with and without experience replay across all training stages. When applying experience replay, we use a subsampling ratio of $s=0.005$. For the initial dataset $\mathcal{D}_0$, the LLM’s original training data, we select a text instruction-tuning dataset generated by the same LLM. We assume that its distribution closely approximates that of the original training data.

\textbf{Mixed strategy:} In our evaluation, we also apply additional mitigation strategies to models trained with experience replay, resulting in two additional baseline settings: ``model merging after experience replay'' and ``discounting the LoRA scaling factor after experience replay''. All mitigation strategy parameters remain consistent with the previous settings.

\begin{table*}[t]\small 
\setlength\tabcolsep{3pt} 
\renewcommand{\arraystretch}{0.9}
\caption{Accuracies (\%) for various strategies to mitigate catastrophic forgetting. \textbf{Merge}: Model merging. \textbf{Scaling}: Scaling the LoRA factor. \textbf{w/ R}: With Experience Replay. The results are reported across four datasets: LLaMA, Web, Trivia, and IFEval, with each dataset further evaluated on different tasks (T2T, S2T, S2S).}

\centering

\begin{tabular}{l|ccc|ccc|ccc|cc} 
  \toprule
  \multirow{3}{*}{\textbf{Mitigation Strategy}} & \multicolumn{3}{c|}{\textbf{LLaMA}} & \multicolumn{3}{c|}{\textbf{Web}} & \multicolumn{3}{c|}{\textbf{Trivia}} & \multicolumn{2}{c}{\textbf{IFEval}} \\ 
  \cmidrule(lr){2-12}

    & \multirow{2}{*}{T2T} & \multirow{2}{*}{S2T} & \multirow{2}{*}{S2S} & \multirow{2}{*}{T2T} & \multirow{2}{*}{S2T} & \multirow{2}{*}{S2S} & \multirow{2}{*}{T2T} & \multirow{2}{*}{S2T} & \multirow{2}{*}{S2S} & \multicolumn{2}{c}{T2T} \\
   & & & & & & & & & & Prompt & Instruction \\ 
  \midrule
    Original & 70.0 & - & - & 61.5 & - & - & 78.9 & - & - & 67.1 & 77.1 \\
    \midrule
    None & 14.3 & 7.3 & \textbf{8.0} & 3.6 & \textbf{1.5} & 0.8 & 6.0 & 3.1 & 1.9 & 9.2 & 20.1 \\
    Merge (Linear) & \textbf{19.3} & \textbf{9.0} & 7.7 & \textbf{5.6} & 1.1 & 0.6 & \textbf{7.8} & \textbf{3.9} & 1.1 & 8.5 & 19.3 \\
    Merge (TIES) & 12.7 & 6.3 & 2.7 & 3.7 & 0.7 & 0.0 & 4.9 & 2.0 & 0.4 & 10.9 & 22.2 \\
    Merge (DARE) & 4.0 & 6.0 & 1.7 & 1.2 & 0.7 & 0.0 & 1.3 & 2.0 & 0.3 & \textbf{11.8} & \textbf{23.1} \\
    Scaling ($\alpha=15$) & 15.0 & 7.0 & 6.7 & 4.5 & 1.3 & \textbf{1.1} & 5.9 & \textbf{3.9} & \textbf{2.3} & 9.1 & 18.6 \\
    Scaling ($\alpha=14$) & 16.0 & 7.3 & 6.7 & 4.0 & 1.2 & 0.3 & 0.3 & 3.3 & 1.9 & 7.9 & 18.3 \\
    \midrule
    Replay  & 66.3 & 50.3 & \textbf{28.7} & 55.2 & 24.2 & \textbf{9.1} & 66.4 & 25.2 & 11.1 & 47.5 & 57.9 \\
    Merge (Linear) w/ R & 68.0 & 44.7 & 16.7 & \textbf{56.4} & 19.2 & 3.6 & \textbf{68.8} & 16.3 & 5.2 & \textbf{50.3} & \textbf{61.3} \\
    Merge (TIES) w/ R & 66.7 & 42.0 & 17.0 & 55.3 & 17.6 & 2.7 & 67.5 & 15.9 & 4.8 & 50.1 & 60.0 \\
    Merge (DARE) w/ R & \textbf{69.3} & 40.0 & 15.7 & 53.1 & 18.4 & 3.0 & 64.0 & 14.3 & 3.8 & 43.8 & 52.2 \\
    Scaling ($\alpha=15$) w/ R & 68.7 & \textbf{52.7} & \textbf{28.7} & 54.4 & 24.9 & 8.0 & 68.6 & 25.4 & \textbf{11.2} & 43.7 & 59.4 \\
    Scaling ($\alpha=14$) w/ R & 68.7 & 50.7 & \textbf{28.7} & 55.3 & \textbf{25.5} & 6.9 & 66.5 & \textbf{27.2} & 11.1 & 49.0 & 60.6 \\

  \bottomrule

\end{tabular}
\label{tab:main_result}
\end{table*}
\subsection{Datasets}

This section describes datasets, benchmarks, and their preprocessing methods used for training and evaluation.

\subsubsection{Training} We use LibriSpeech 960-hour\cite{7178964} for ASR and TTS training and Magpie-Air \cite{xu2024magpie}\footnote{Magpie-Align/Llama-3-Magpie-Air-3M-v0.1} for SQA and experience replay.
    
\textbf{In ASR stage,} LibriSpeech is used for training, with each example assigned one of ten randomly selected instructions, such as "Please repeat the following words:".

\textbf{In TTS stage,} LibriSpeech is also used, with randomly assigned instructions like "Please speak out loud the following words:". To generate word-by-word text-speech interleaved sequences, we align text transcriptions and speech utterances at the word level using Whisper-Timestamped\footnote{https://github.com/linto-ai/whisper-timestamped}.

\textbf{In SQA stage,} Magpie-Air is used for training. We filter examples based on length and topic, removing categories like math and coding that are challenging to describe in speech. Speech questions and answers are synthesized from text using SpeechT5~\cite{ao-etal-2022-speecht5}, creating both text and speech versions of QA pairs. Whisper-Timestamped is used to align answers and generate interleaved text-speech token sequences.

When applying experience replay, we use Magpie-Air as $\mathcal{D}_0$, randomly sampling text instruction-response pairs for training. Since Magpie-Air is constructed by prompting Llama 3 8B, an LLM from the same series as ours, it serves as a suitable dataset for experience replay.

\subsubsection{Evaluation} 
To assess the model's learned capabilities, we evaluate spoken question answering (SQA) on the speech modality. To measure the catastrophic forgetting in the SLM, we evaluate question answering (QA) and instruction-following on the text modality.

\textbf{Spoken question answering:} Following the evaluation settings of Moshi and Spectron~\cite{nachmani2023spoken}, we use three datasets for SQA: (1) Spoken WebQuestions, (2) LLaMA-Questions and (3) Audio Trivia QA. SQA is evaluated under two settings: speech-to-text (S2T), where accuracy is computed by directly matching text responses, and speech-to-speech (S2S), where the speech responses is first transcribed with Whisper-large-v3 and considered correct if the transcription contains the correct answer.

\textbf{Question answering:} We use the same datasets as in SQA but evaluate only on the text modality (Text-to-Text, T2T). Accuracy is used as the evaluation metric.

\textbf{Instruction following:} We use IFEval~\cite{zhou2023instruction} for evaluating instruction-following on the text modality (T2T setting). IFEval computes accuracy at both the prompt and instruction level to assess the model’s ability to follow instructions accurately.


\begin{figure}[t]
  \centering
  \includegraphics[width=\linewidth]{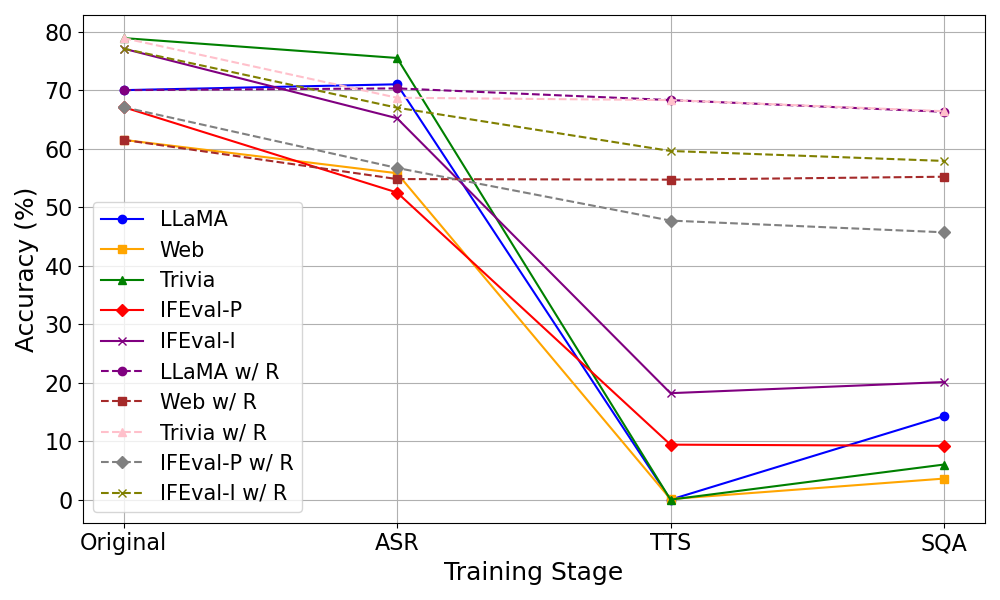}
  \caption{Evaluation results on instruction-following and question answering. LLaMA, Web, and Trivia denote LLaMA-Questions, Spoken WebQuestions, and Audio Trivia QA. IFEval-P and IFEval-I stand for IFEval in prompt-level and instruction-level. w/ R means with experience replay.}
  \label{fig:train}
  \vspace{-10pt}
\end{figure}

\section{Results}

\subsection{Catastrophic forgetting}
Fig.\ref{fig:train} shows the evaluation results on instruction-following and question answering in each training stage on T2T setting. For SLM without any mitigation strategy, it is obvious that catastrophic forgetting appear during training. As training stage moves on, the accuracy of both evaluation tasks decrease in different degrees. There is a gap can be observed easily between ASR stage and TTS stage on both evaluation tasks, which also shows that the most serious forgetting appears in TTS stage. Interestingly, accuracy for question anawering slightly grow in SQA stage. We assume this is because the SLM recalls some of its knowledge from the text sequence at this stage, even if only to a small extent. As for SLM applied with experience replay, although there is still a little forgetting during training, the extent has been mitigated significantly compared to one without experience replay. 

\subsection{Mitigation strategies}
Table.\ref{tab:main_result} shows the results for all mitigation strategies. From the results, the are several findings:

\textbf{(1) Experience replay surpasses the other strategies:} According to the results, experience replay surpasses all the other single strategies on tasks evaluated for mitigation (T2T) as well as new ability (S2T, S2S).

\textbf{(2) Mixed Strategy can further boost the performance:} Compared to experience replay, other mixed strategies can achieve better performance of new ability in S2T setting in some cases. However, mixed strategy with discounting LoRA-scaling factor more robust than model merging.

\textbf{(3) Experience replay remains robustness in every setting:} Although all mitigation strategies can do some improvements in almost all settings. However, only experience replay remains robust in S2S setting and  surpasses other strategies.

\section{Conclusion}
This paper investigates mitigation strategies for continual learning in developing spoken language models (SLMs) from large language models (LLMs). The results demonstrate that experience replay is the most effective method, with further performance gains achievable by combining it with other techniques. Through a case study, we highlight catastrophic forgetting as a significant challenge and showcase the potential of these strategies to address it. Future work will involve more comprehensive studies, including diverse training pipelines, models and various strategies to inspire the speech community to develop more efficient training methods.
\bibliographystyle{IEEEtran}
\bibliography{mybib}

\end{document}